\pdfoutput=1
\PassOptionsToPackage{dvipsnames, svgnames, x11names}{xcolor} 
\documentclass[11pt]{article}

\usepackage[preprint]{acl}

\usepackage{times}
\usepackage{latexsym}
\usepackage{booktabs}
\usepackage[T1]{fontenc}

\usepackage[utf8]{inputenc}

\usepackage{microtype}

\usepackage{inconsolata}

\usepackage{amsmath} 
\usepackage{array} 
\usepackage{multirow}
\usepackage{graphicx}
\usepackage{microtype}
\usepackage{inconsolata}

\usepackage{times}
\usepackage{latexsym}
\usepackage{adjustbox} 
\usepackage{graphicx}
\usepackage{listings}
\usepackage{todonotes}
\usepackage{xcolor}
\usepackage[most]{tcolorbox}
\usepackage{tocloft}
\definecolor{promptbox_color}{HTML}{9DC8C8}
\definecolor{promptbox_color_2}{HTML}{a5d296}

\usepackage{url}

\usepackage{xcolor}
\usepackage{enumitem}
\PassOptionsToPackage{dvipsnames, svgnames, x11names}{xcolor}
\usepackage{tcolorbox} 

\title{Too Open for Opinion? Embracing Open-Endedness in \\Large Language Models for Social Simulation}

\author{
  \textbf{Bolei Ma\textsuperscript{1}}~~~
  \textbf{Yong Cao\textsuperscript{2}}~~~
    \textbf{Indira Sen\textsuperscript{3}}~~~
  \textbf{Anna-Carolina Haensch\textsuperscript{1,4}}~~~
  \\\vspace{5pt}
\textbf{Frauke Kreuter\textsuperscript{1,4}}~~~
\textbf{Barbara Plank\textsuperscript{1}}~~~
  \textbf{Daniel Hershcovich\textsuperscript{5}}
\\
  \textsuperscript{1}LMU Munich \& Munich Center for Machine Learning, \\
  \textsuperscript{2}University of Tübingen \& Tübingen AI Center, 
    \textsuperscript{3}University of Mannheim, \\\vspace{5pt}
  \textsuperscript{4}University of Maryland, College Park, 
  \textsuperscript{5}University of Copenhagen 
\\
    \texttt{bolei.ma@lmu.de, yong.cao@uni-tuebingen.de, dh@di.ku.dk}
}
\begin{document}
\maketitle

\begin{abstract}

Large Language Models (LLMs) are increasingly used to simulate public opinion and other social phenomena. Most current studies constrain these simulations to multiple-choice or short-answer formats for ease of scoring and comparison, but such closed designs overlook the inherently generative nature of LLMs. In this position paper, we argue that open-endedness, using free-form text that captures topics,  viewpoints, and reasoning processes ``in'' LLMs, is essential for realistic social simulation. 
Drawing on decades of survey methodology research and recent advances in NLP, we argue why this open-endedness is valuable in LLM social simulations, showing how it can improve \emph{measurement} and \emph{design}, support \emph{exploration} of unanticipated views, and reduce researcher-imposed \emph{directive bias}. It also captures \emph{expressiveness} and \emph{individuality}, aids in \emph{pretesting}, and ultimately enhances \emph{methodological utility}. 
We call for novel practices and evaluation frameworks that leverage rather than constrain the open-ended generative diversity of LLMs, creating synergies between NLP and social science.

\end{abstract}

\section{Introduction}

Since NLP technologies are increasingly used in social situations, recognizing the importance of social context has become critical, and LLMs are now being applied to socially aware tasks \cite{hovy-yang-2021-importance,ziems-etal-2024-large,yang-etal-2025-socially}.  
A prominent example is social simulation, where synthetic agents are used to explore collective attitudes, policy preferences, and cultural dynamics (\citealp[][\emph{inter alia}]{santurkar2023whose,rottger-etal-2024-political,ma2024potential,anthis2025position,Kozlowski2025}).  
A typical setup starts by prompting an LLM to ``play'' the role of virtual personas, conditioned on participants' profiles or past behavior data. The researcher then aggregates their responses 
for example,  
to track shifting public opinion or simulate human behaviors. 
Recent work highlights the promise of such simulations for generating timely, low-cost social data and for testing sociological theories at scale \cite{anthis2025position,wang2025limitsllmbasedhumansimulation,li2025positionsimulatingsocietyrequires}. This potential has sparked burgeoning interest within the NLP community and beyond.\footnote{See, for example, this First Workshop on Social Simulation with LLMs at COLM'25: \url{https://sites.google.com/view/social-sims-with-llms/}.} 

The majority of the most current studies usually adopt a survey-like multiple-choice or short-answer format to keep model outputs easy to score and compare \cite{meister-etal-2025-benchmarking,balepur-etal-2025-best,ni2025surveylargelanguagemodel}.  
While convenient, such closed forms risk collapsing nuanced opinions, steering responses toward researcher-defined categories, and masking the generative strengths of the LLMs \cite{lyu-etal-2024-beyond,wang2024look,wang2024my,wang-etal-2025-ubench}.  

\begin{figure}[t]
    \centering
    \includegraphics[width=1\linewidth]{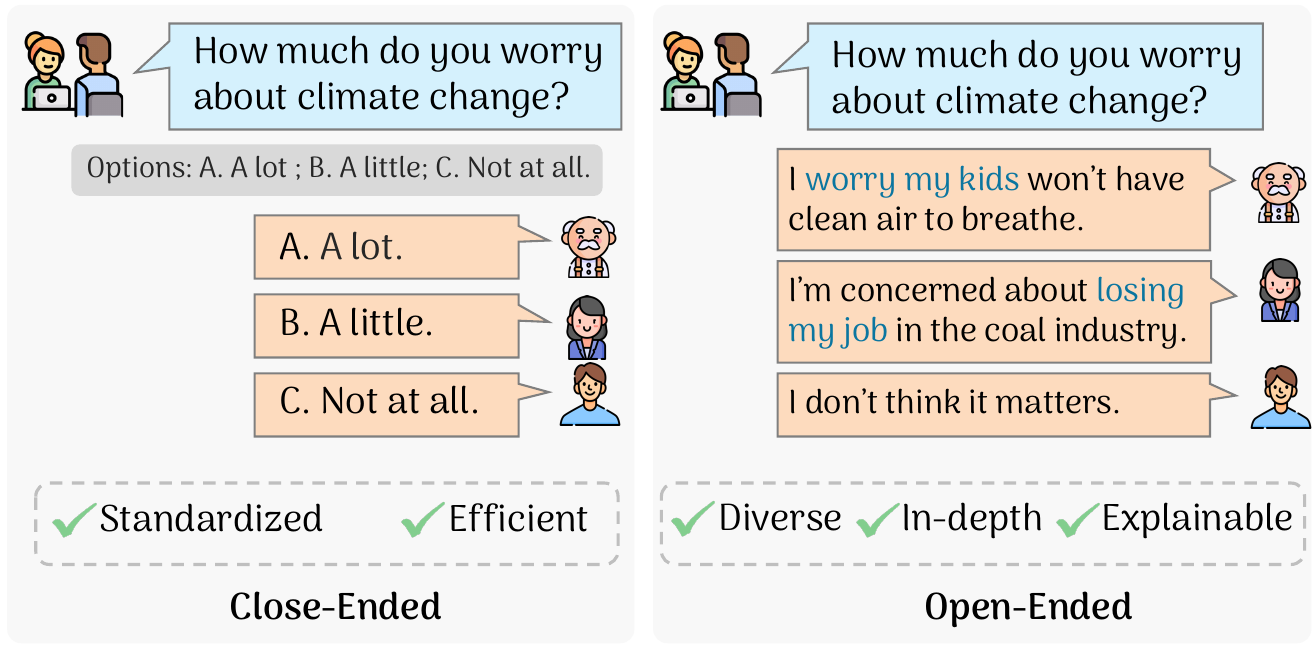}
    \caption{Example of close-ended and open-ended surveying, illustrating that open-ended interaction better captures participant intentions, generates more diverse and explainable responses, and enables more accurate social simulation compared to a closed-ended one. }
    \label{fig:close-open}
\end{figure}

This position paper argues that social simulations should instead embrace open-endedness: the capacity of LLMs to produce free text that is as varied and unpredictable as human discourse.  
Open-ended questions have long been central to \emph{survey methodology}\footnote{The term \emph{survey} in survey methodology research refers to a systematic method for gathering information from a sample of a population, different from the sense used for literature review in the NLP/ML community. 
Survey methodology is a well-established discipline with its own principles, theories, and scientific standards in social science \cite{groves2009}.} because they surface unanticipated topics, capture minority viewpoints, and reveal reasoning processes that fixed options miss \cite{Zuell2016,Singer2017}.  
Recent scholarship further underscores the deep parallels between NLP and survey research, noting shared methodological goals and opportunities for cross-fertilization \cite{kern-etal-2023-annotation,pmlr-v235-eckman24a,sen2025connecting}.\footnote{See, also, this First Workshop on Bridging NLP and Public Opinion Research at COLM'25: \url{https://sites.google.com/view/nlpor2025}.}  
These same properties are vital for simulations meant to approximate real social complexity.  
Thus, insights from decades of social science research on eliciting, coding\footnote{In social science, \emph{coding} refers to annotating free responses into analytic categories \cite{he2020automatic}, distinct from the meaning of ``programming’’ in NLP/ML.}, and interpreting open responses  offer practical guidance for handling the variability of LLM outputs (\citealp[e.g.,][]{Schonlau2016,haensch2022semi,Gweon2023}).

\begin{tcolorbox}[
    colframe=promptbox_color,      
    colback=white,                 
    coltitle=black,                
    colbacktitle=promptbox_color,  
    rounded corners,
    arc=1.5mm, 
    enhanced,                    
    boxrule=0.6mm,                
    frame style={solid},          
    fonttitle=\bfseries,          
    fontupper=\normalsize,
    title={Position: Embracing Open-Endedness.}
]
Embracing open-ended generation in social simulation allows researchers to capture heterogeneity in opinions, uncover minority viewpoints, and examine reasoning patterns, all of which are central to realistic social modeling.

\end{tcolorbox}

Our paper is organized as follows:  
§\ref{sec:open-endedness} introduces the concept of open-endedness in LLM social simulation.  
§\ref{sec:current} reviews current practices on close-ends and limitations.  
§\ref{sec:arguments} presents our main arguments for embracing open-endedness, drawing on survey methodology and promising use-cases.  
§\ref{sec:practical} offers methodological insights from social science and NLP for analyzing open-ended LLM outputs. §\ref{sec:challenges} summarizes major challenges and future directions.

\section{Open-Endedness in LLM Social Simulation}
\label{sec:open-endedness}

What is \emph{open-endedness} or, more specifically, what are \emph{open-ended questions} in surveys? Compared to closed-ended questions, where a fixed set of selectable response options is provided to participants, ``all survey questions that do not include a set of response options are known as open-ended questions'' \cite{Zuell2016}. Their defining characteristic is that respondents are free to formulate answers in their own words, thereby introducing greater variability and richness into the collected data \cite{ca47b63b-bde5-3ed2-b105-ccdface181f8, krosnick1999survey,bradburn2004asking,KrosnickPresser2010,dillman2014tailored}. 
Figure~\ref{fig:close-open} illustrates a typical closed-ended question (left) versus an open-ended question (right). 

By analogy, \emph{open-endedness in LLMs}\footnote{In this paper, we refer to \emph{open-endedness}, \emph{open-end} and \emph{free-form generation} in LLMs as synonyms.} refers to the capacity of these models to generate free-form, unconstrained text in response to a prompt, rather than being restricted to a predefined set of options. This property emerges directly from the fact they are language models, that is, they model the probability distribution over token sequences. LLMs predict the next token based on a distribution over possible continuations, without being inherently tied to a fixed response space. 
It is important, however, to distinguish between \emph{open-ended generation} and \emph{closed-ended tasks} in LLMs. 
Even when prompting with closed-ended questions, researchers still use models to generate text autoregressively, and extract categorical answers by mapping the generated text back to predefined labels (e.g., via string matching in \citealp{argyle-etal-2023,vonderheyde2024vox}) or by using token-level probabilities over constrained options (\citealp[see e.g.,][]{santurkar2023whose, shu-etal-2024-dont, cao-etal-2025-specializing}).

The crucial distinction, therefore, lies not in the generation process itself, but in the \textbf{structure of the task input design}: open-ended prompts leave the model unconstrained, whereas closed-ended prompts channel output into researcher-defined categories. For this work, we define \emph{open-endedness} as the use of prompts that elicit unconstrained, free-form responses from LLMs without restricting outputs to a fixed set of options.

In NLP, \emph{open-ended generation} typically refers to tasks where models are valued for creativity and diversity
\cite{Holtzman2020The}.\footnote{Such as story writing
\cite{fan-etal-2018-hierarchical}, open-domain dialogue
\cite{zhang-etal-2020-dialogpt}, or long-form question answering
\cite{krishna-etal-2021-hurdles}.} 
Our focus goes beyond: we treat open-endedness not only as artistic free play or design choice but also as a \textbf{methodological contribution for social simulation}, where the goal is to approximate the diversity of real human opinions and reasoning in addition to maximizing novelty or narrative flair. 
Here, simulation does not mean producing a single representative answer but generating multiple, potentially divergent responses that together approximate a range of plausible human perspectives \cite{anthis2025position}, and ideally including cognitively grounded reasoning \cite{li2025positionsimulatingsocietyrequires}. In this sense, open-ended generation is not only a technical feature of these models but also a defining characteristic of their emerging role in simulating social processes and communicative variability.

\section{Current Practice and Limitations on Close-Ends in LLM Social Simulation}
\label{sec:current}

Despite the generative capabilities described in §\ref{sec:open-endedness},
most LLM-based social simulations continue to favor \emph{closed-ended} designs. From a follow-up review from the 53 studies cataloged in \citet{anthis2025position}, we find that only 11 (21\%) include any open-text component, and only 4 (8\%) rely primarily on free-form outputs during evaluation. Similarly, in a systematic review of LLM-based social simulations,~\citet{sen-etal-2025-missing} find that a majority of studies use closed-ended response formats. In short, the majority of current simulations reproduce the logic of traditional survey instruments, in
which models (like human respondents) are constrained to predefined response options.

This reliance on closed-ended designs for LLM-based social simulation is understandable. Multiple-choice and categorical response formats enable straightforward aggregation, facilitate quantitative comparison, and align with long-established practices in survey research \cite[\emph{inter alia}]{argyle-etal-2023,santurkar2023whose,durmus2024towards}. For tasks such as  measuring opinions or eliciting stated preferences from simulated agents, these formats provide scalability at low cost.

Yet these same conveniences restrict what makes LLMs distinctive.  
Constraining outputs to predefined labels suppresses the unexpected topics, minority viewpoints, and context-sensitive reasoning that open-ended generation can reveal \cite{Schonlau2016,Gweon2023}.  
It can also introduce \emph{aggregation bias}, where nuanced opinions must be forced into overly broad categories \citep{Tijdens2014}, and risks ``straight-lining’’ or superficial responses that mask underlying attitudes.  
Recent critiques of multiple-choice evaluation in NLP echo these concerns, noting that fixed options reward recall over reasoning and may hide model biases or unfaithful explanations \citep{balepur-etal-2025-best}.  
The result is a simulation that reproduces the limitations of traditional surveys rather than exploiting the richer variability of generative models.

Current practice, therefore, reveals a tension. Closed-ended tasks remain convenient, but they underuse the very property, generative openness, which distinguishes LLMs from earlier computational agents. If the goal of social simulation is to capture the diversity and complexity of human attitudes, open-ended responses should be treated not as noise to be filtered out but as a core signal to be analyzed. 
By drawing on the extensive literature on open-ended survey questions and LLM capabilities, we can design prompts and simulations that preserve expressive richness while supporting downstream analysis. We  
detail the benefits in §\ref{sec:arguments}.

\begin{table*}[htbp]
\centering
\renewcommand\arraystretch{1.5}
\small
\begin{tabular}{p{2.2cm}|p{6cm} p{6.5cm}}
\toprule
\textbf{Dimension} & \textbf{Human Surveys (Social Science Domain)} & \textbf{LLM Social Simulation (NLP Domain)} \\
\midrule \midrule
Measurement &
Improves validity and reduces guessing. &
Open text generation provides reasoning chains and internal consistency checks. \\
\midrule
Exploration &
Reveals unexpected or minority views. &
Surfaces emergent opinions beyond predefined categories. \\
\midrule
Design &
Avoids long option lists or forced categories. &
Handles large answer spaces with minimal prompt constraints. \\
\midrule
Directive Bias &
Reduces researcher-induced steering. &
Allows diverse framings, lowering prompt or category bias. \\
\midrule
Pretesting &
Shows how respondents interpret questions. &
Helps diagnose prompt effects and refine the simulations. \\
\midrule
Engagement \& Expressiveness &
Encourages motivation and engagement with richer, more detailed articulation. &
Generates varied, more expressive human-like responses beyond standardized replies. \\
\midrule
Individuality &
Captures unique, personalized phrasing. &
Supports heterogeneous synthetic populations and natural variation. \\
\midrule
Methodological Utility &
Offers multi-perspective material for analysis. &
Serves as “tools” and alternative framings that enrich downstream analyses. \\
\bottomrule
\end{tabular}
\caption{Benefits of open-ended questions in survey research (social science) and their implications for LLM-based social simulation (NLP), highlighting how open-ended responses enhanced corresponding tasks. }
\label{tab:open_ends}
\end{table*}

\section{Arguments for Open-Endedness}
\label{sec:arguments}
Table~\ref{tab:open_ends} summarizes established benefits of open-ended questions in human survey methodology and maps them to their potential implications for LLM-based social simulation. In this section, we discuss these parallels in more detail, in order to provide arguments for why open-ends are important in LLM social simulations. We begin by reviewing insights from survey research (§\ref{sec:lessons human studies}), then extend these lessons to LLM social simulations (§\ref{sec:implications llm socialsim}) to support our key position, and finally outline promising application areas (§\ref{sec:usecases}).

\subsection{Lessons from Human Studies}
\label{sec:lessons human studies}

Survey methodology has long emphasized the complementary role of open-ended questions alongside closed-ended ones. Decades of research show that open formats allow researchers to capture knowledge and attitudes in a way that structured response options cannot (\citealp[e.g.,][]{ca47b63b-bde5-3ed2-b105-ccdface181f8,krosnick1999survey,bradburn2004asking,groves2009,KrosnickPresser2010,Zuell2016,Albudaiwi2017SurveyOpenEnded}). Drawing from these advances in survey research, we summarize the lessons and benefits in eight dimensions. 

\paragraph{Measurement.} Open-ended formats often yield more accurate knowledge measurement \cite{geer1988openended,Zuell2016}.  Respondents’ answers can better reflect what they genuinely know or recall. Although they can sometimes increase “don’t know” or refusal rates, such responses themselves provide useful diagnostic information about the limits of respondents’ recall or confidence \cite{KrosnickPresser2010,Albudaiwi2017SurveyOpenEnded} and the reasons for refusal \cite{Singer2017}.

\paragraph{Exploration.} Open-ended questions are particularly useful when the full range of possible answers is unknown. They can surface new, unexpected, emergent, or rare perspectives that researchers might not anticipate in advance \cite{Zuell2016,Albudaiwi2017SurveyOpenEnded}. They can help avoid priming preset answers,  eliciting ``first-order'' thinking. \cite{ferrario2022eliciting}. 
For example, in attitude surveys, respondents may highlight concerns that fall outside predefined categories, offering insights into emergent public issues.  

\paragraph{Design.} Open-ended formats avoid excessively long lists of response options, which can overwhelm respondents when the answer space is very large, such as when classifying occupations or consumer products \cite{Zuell2016}. Rather than presenting exhaustive lists, open-ended questions provide a direct way to capture variation more naturally, reducing the effort required in option design.

\paragraph{Directive Bias.} Open-ended designs help reduce directive bias, which is also called ``framing effect'', where the way how a choice is presented can lead to selection bias \cite{tversky1981framing, schuman1996questions, tourangeau2000psychology}. By not steering respondents toward predefined categories, they ensure that answers reflect genuine perspectives rather than artifacts of survey design \cite{Zuell2016}. This is particularly important in sensitive domains \cite{Albudaiwi2017SurveyOpenEnded}, where question wording and free-from response framing can shape reported attitudes. 
They help avoid prejudging respondents \cite{schuldt2014media}. 

\paragraph{Pretesting.}  Open-ended questions support cognitive pretesting. They provide insights into how respondents interpret the meaning of questions and reveal the reasoning behind their choices \cite{Lenzner2015KognitivesPretesting}
. Responses to the open-ended probes provide vital information on respondents’ potential need for clarification and how to improve the survey questions \cite{Singer2017,Neuert2021}. This can be invaluable during questionnaire design, as it uncovers misunderstandings or unintended interpretations that would remain hidden in closed formats.  

\paragraph{Engagement and Expressiveness.} Open-ended questions can enhance respondent engagement, giving respondents a word \cite{Singer2017}. Strategically placed open-text prompts give participants space to articulate thoughts in their own words, often inviting them to share personal stories or express hesitation, reluctance, or emotional nuance that closed formats cannot capture \cite{Albudaiwi2017SurveyOpenEnded}, usually at the end of an attitude survey \cite{Porst2014}. This sense of involvement enriches the data and can improve quality of responses \cite{ConnorDesai2019OpenClosedCIE}.

\paragraph{Individuality.} Open-ended formats capture individuality and natural variation. Each respondent’s articulation is personalized with a sense of individuality, unique and sometimes creative, reflecting their particular circumstances or expressive style \cite{Albudaiwi2017SurveyOpenEnded}. This variability, though more difficult to compare systematically, is valuable: it produces data that mirrors the diversity and naturalness of real social phenomena.  

\paragraph{Methodological Utility.} Open-ended responses serve as valuable qualitative methodological resources beyond raw data contribution \cite{Singer2017}. Researchers can directly quote them to support findings, use them to identify relevant vocabulary, or triangulate across different perspectives on a topic. They allow studies not only to present statistical results but also to incorporate authentic ``voices'' to the research methodology.

\subsection{Implications for LLM Social Simulations}
\label{sec:implications llm socialsim}

These social science lessons carry useful implications for LLM social simulations. Open-ended responses leverage what distinguishes LLMs from previous computational approaches: their ability to generate free-form, context-sensitive text. We now draw on the implications based on findings in social science for constructing arguments for the use of open-ends if LLMs are used for social simulation, also shown in Table \ref{tab:open_ends}.

\paragraph{Measurement.} Open-ended LLM responses often include justifications or contextual details, allowing researchers to assess the plausibility and internal consistency of the synthetic output. This helps determine whether the model has captured the simulated persona. 
For example, a conservative respondent might explain skepticism about climate change by citing economic priorities and shifting scientific claims, through an open-ended reasoning and explanation. Such reasoning, absent in closed-ended responses, 
mirrors the broader strategy of using a model's elicited reasoning steps as a way to validate it self's conclusions \cite{wei2022chain}.

\paragraph{Exploration.} LLMs are capable of surfacing perspectives that are not anticipated in advance \cite{ma2024potential}. By using open-ended tasks, researchers can elicit emergent themes, viewpoints of hard to reach subpopulations, and unstructured arguments that mirror the unpredictability of real populations. This is especially valuable in contexts where survey research struggles, such as simulating hard-to-reach or underrepresented groups \cite{vonderheyde2024vox,namikoshi2024using}. Instead of forcing all responses into predefined categories, open-ended generation allows for the discovery of attitudes and framings that would otherwise remain invisible. Such novel and complex outcomes in AI could only realized through open-ended generation \cite{stanley2015greatness}.

\paragraph{Design.} Traditional surveys often face design bottlenecks when the possible answer space is large or complex \cite{bradburn2004asking,KrosnickPresser2010}. In contrast, LLMs can easily generate responses across vast domains including political justifications, occupational identities, cultural references, without requiring the researcher to enumerate all possible categories in advance \cite{10.1145/3586183.3606763}. This not only reduces the design burden but also leads to more realistic variation in the simulated population, reflecting the breadth of how humans articulate similar underlying attitudes in many different ways.

\paragraph{Directive Bias.} LLM outputs can be affected by how researchers prompt them \cite{schulhoff2025prompt,bardol2025chatgptreadstoneresponds}. Closed-ended questions inevitably steer models toward predefined frames, mirroring the same risk of directive bias in human surveys. 
The choice of answer options and how it is presented can both affect the final results \cite{zheng2024large,pezeshkpour-hruschka-2024-large,wei-etal-2024-unveiling}. 
Open-ended prompts mitigate this risk by allowing models to generate their own framings and vocabularies. For instance, when asked about climate change or political attitudes, an LLM might emphasize scientific uncertainty, intergenerational justice, or geopolitical responsibility depending on the simulated persona, rather than being constrained to ``agree/disagree'' scales. This autonomy better approximates natural opinion expression and reduces researcher-induced biases in LLM outputs.

\paragraph{Pretesting.}  
Just as open-ended human responses are useful for testing whether survey items are well understood, open-ended LLM responses can be used to probe how prompts are ``interpreted''. By inspecting generated rationales, researchers can diagnose whether a prompt unintentionally suggests an interpretation or introduces ambiguity. This supports iterative refinement of both survey instruments and simulation protocols, ensuring greater robustness. In effect, LLMs can be deployed as large-scale cognitive pretesters/advisors/pilot testers \cite{kim2024llm,adhikari2025exploringllmsautomatedgeneration,beck2025biasloophumansevaluate,Cui2025}, helping with both real human / LLM full experiments.   

\paragraph{Engagement and Expressiveness.}  
Although LLMs do not face motivational constraints, open-ended generation enriches the expressive quality of simulated responses. Instead of one-dimensional choices, outputs may include narratives, hesitations, or rhetorical flourishes, which create more human-like populations. Such expressive variation is not merely decorative: it allows simulations to capture how attitudes are embedded in language practices such as sarcasm, emotional appeals, or careful hedging, adding a qualitative layer of realism that purely quantitative approaches lack.  

\paragraph{Individuality.}  
One of the most valuable features of open-ended LLM outputs is their ability to simulate heterogeneity. Rather than collapsing responses into standardized categories, models can generate diverse, idiosyncratic articulations of similar attitudes. This variability mirrors the individuality observed in real populations, where two people with the “same” opinion might justify it in strikingly different ways. Incorporating such micro-level differences supports the construction of synthetic agents or societies that are not only statistically representative but also qualitatively rich \cite{park_Social_Simulacra, park2024generative}.  

\paragraph{Methodological Utility.}  
Open-ended LLM outputs provide not only data but also analytic resources. Just as survey researchers quote respondents or analyze vocabularies, LLM researchers can use LLM-generated synthetic texts as ``tools'' that illustrate findings, extract salient terms to map semantic fields, or triangulate different perspectives on an issue \cite{liu2024best,long-etal-2024-llms,park2024generative}. In this sense, open-ended simulation may not merely reproduce distributions of attitudes but also provides 
interpretable and quotable material, enhancing both computational and qualitative analysis of synthetic societies.  

\subsection{Use-Cases}
\label{sec:usecases}

Based on these benefits, we outline promising (but not exhaustive) use-cases of open-ended simulations.  
The detailed use-cases and implications are summarized in Appendix §\ref{sec:datailed use-cases}.

\paragraph{Populations and opinion simulation.} By conditioning on demographic, ideological, or cultural profiles, LLMs can approximate subgroup-specific “silicon samples”. Crucially, open-ended responses capture not only preferences but also the rationales and nuances often missed in closed-choice polls. While current LLMs may not fully replicate human behavior \cite{Chemero2023LLM,Zhang2025Generative}, studying open-ended outputs is crucial for improving realism and diversity in synthetic populations.

\paragraph{Democratic proxy citizens.} LLMs are increasingly framed as proxy citizens in civic processes \cite{Goni2025CitizenParticipation}. Such proxies offer policymakers dynamic, fine-grained insights, where the open-ended format enables evolving concerns and nuanced justifications beyond fixed-choice surveys. Despite existing gaps, analyzing free-form reasoning helps refine both model alignment and survey design.

\paragraph{Qualitative interviews.} 
LLMs can emulate participants in qualitative research, generating free-form narratives in interviews and focus groups. 
These simulations can reveal emergent themes and discursive patterns. Current models often struggle with coherence and authenticity in extended qualitative contexts in LLM-based interviews \cite{Kapania2025}, but open-ended prompts provide a framework to 
study and improve these abilities.

\paragraph{Deliberation and debate simulations.} 
Open-ended prompting allows LLMs to generate arguments, counterarguments, and rhetorical strategies that closed-choice setups cannot anticipate. LLMs can therefore be useful for modeling communicative dynamics, especially in multi-agent debates.

\paragraph{Behavior trials.}  In behavioral experiments, open-ended LLM responses yield richer data than fixed-choice outcomes by revealing reasoning and strategies. 
While LLMs cannot fully replace human participants, studying their open-ended behavior helps refine experimental design and theory testing before deployment.

\paragraph{Social media use-cases.} LLMs can model large-scale human behavior on digital platforms. 
Though current models might fall short of capturing complex social dynamics \cite{piao2025agent}, open-ended outputs allows agents to display diverse, context-sensitive behaviors.

\paragraph{Model training and evaluation.} 
Open-ended responses are valuable not only for social science but also for advancing NLP methodology itself. We argue that social science methods can shape model alignment and evaluation. Simulated open-ended judgments yield richer signals than binary or scalar preferences, enhancing training and evaluation by revealing subtle reasoning distinctions.

\section{Practical Insights from Social Science and NLP}
\label{sec:practical}

Now, our mission is to design a concrete methodological outlook for studying the ``open-ends'' in social simulations. Decades of social-science practice, from meticulous qualitative coding to today’s state-of-the-art language models, offer lessons for researchers who build or evaluate LLM-based simulations. 
We highlight how insights move in both directions: traditional social science methods such as qualitative coding and content analysis can guide the design and assessment of open-ended simulations, while advanced NLP methods including statistical machine learning workflow, LLM-driven techniques, and emerging hybrid methods can, in turn, inform social-science methodology itself, and also expand the toolkit for social simulation.

\subsection[Social Science to NLP]{Social Science $\rightarrow$ NLP}

\paragraph{Manual and qualitative coding.} The foundational approach is manual coding supported by a well-defined codebook and reliability checks.  
Researchers develop conceptual categories, train multiple coders, and use double-coding or adjudication to ensure validity \cite{Ongena2006,doi:10.1515/jos-2016-0003,Singer2017,he2020automatic,Neuert2021}.  
An example of the coding scheme is shown in Appendix §\ref{sec:coding-scheme}. These practices embody key social science principles, reflexivity, transparency, and intercoder reliability, that directly inform LLM simulations: synthetic personas and prompts should likewise be specified, documented, and independently reviewed to guard against hidden bias. 

\paragraph{Quantitative and exploratory content analyses.}
Beyond coding, social scientists have long applied a range of traditional quantitative techniques to open-ended responses, often beginning with careful data elicitation.  
Web surveys, for example, exploit dynamic filtering and autocomplete to guide respondents through vast occupational or categorical taxonomies while preserving the freedom of open text \cite{Tijdens2014}.  
Once collected, responses are frequently subjected to computer-assisted qualitative data analysis \cite{Singer2017} 
to structure the corpus and prepare it for statistical modeling.  
Researchers then employ exploratory text-analytic tools, such as word-frequency visualizations, keyness or distinctiveness measures, and topic modeling, to quantify patterns and highlight salient themes \cite{ferrario2022eliciting,Rouder2021What}.  
These approaches illustrate how open-ended answers can be transformed into analyzable variables without losing their descriptive richness, offering practical lessons for LLM-based social simulation: careful elicitation ensures coverage of complex answer spaces, while quantitative summaries and visualizations provide scalable ways to monitor and interpret large volumes of synthetic text.

\begin{tcolorbox}[
    colframe=promptbox_color_2,      
    colback=white,                 
    coltitle=black,                
    colbacktitle=promptbox_color_2,  
    rounded corners,
    arc=1.5mm, 
    enhanced,                    
    boxrule=0.6mm,                
    frame style={solid},          
    fonttitle=\bfseries,          
    fontupper=\normalsize,
    title={Lesson: Social Science Informs NLP.}
]

Long-standing traditions of qualitative and quantitative analysis of the open-ends in the social sciences provide essential methodological guardrails, ensuring that when applied to NLP and LLM-based simulations, open-ended text is elicited, coded, and interpreted with rigor, transparency, and validity.

\end{tcolorbox}

\subsection[NLP to Social Science]{NLP $\rightarrow$ Social Science}

\paragraph{From statistical and semi-automatic workflows to neural and Transformer-based methods.} Before deep learning, social scientists turned to supervised and semi-supervised classifiers such as SVMs and random forests to scale up coding \cite{he2021coding,he2021a}.  
Semi-automatic workflows combine machine predictions with targeted human checks \cite{Schonlau2016,haensch2022semi}, showing that human-in-the-loop verification can preserve rigor while improving efficiency.  
Recent work demonstrates the strength of transformer encoders like BERT for multi-label and multilingual survey coding \cite{Schonlau2023multi,Gweon2023}.  
This parallels the need to adapt models to particular simulated populations: just as social scientists curate training data that reflect a target electorate, simulation researchers can condition models on carefully chosen community texts or historical corpora. 

\paragraph{LLMs for qualitative coding.} A big methodological benefit is to use LLMs themselves as coders. 
TopicGPT \cite{pham-etal-2024-topicgpt}, and other survey based coding work (\citealp[e.g.][]{Dunivin2025,doi:10.1177/08944393241286471,vonderheyde2025aintsurveyusinglarge}) explore zero- and few-shot coding of open-ended survey responses.  
These findings suggest that, with careful prompting and validation, LLMs can approach the reliability of trained human coders while providing richer reasoning traces.  
For social simulation, this highlights a dual role: LLMs are not only objects of study but also methodological partners for coding, auditing, and refining the very synthetic data they generate.

\paragraph{Hybrid strategies and human–machine collaboration.}
Many contemporary projects integrate human and machine strengths: analysts create the initial scheme, models propose labels, and disagreements are resolved collaboratively \cite{Wilson2022,haensch2022semi}, similar to active learning in ML/NLP, where models and humans iteratively collaborate to improve labeling efficiency and quality \cite{settles2009active,zhang-etal-2022-survey}.  
This iterative interplay mirrors long-standing ethnographic and survey norms of triangulation, combining multiple perspectives to strengthen inference. 
A recent study on German opinions in LLMs \cite{ma-etal-2025-algorithmic} combined human-coded subsets with a German BERT classifier, illustrating effective human–model integration. 
For LLM simulations, hybrid strategies enable cyclical workflows where model outputs inform new simulations, experts provide feedback, and the model is updated or re-prompted; alternatively, experts can pre-annotate data to guide subsequent model updates.

\begin{tcolorbox}[
    colframe=promptbox_color_2,      
    colback=white,                 
    coltitle=black,                
    colbacktitle=promptbox_color_2,  
    rounded corners,
    arc=1.5mm, 
    enhanced,                    
    boxrule=0.6mm,                
    frame style={solid},          
    fonttitle=\bfseries,          
    fontupper=\normalsize,
    title={Lesson: NLP Advances Social Science.}
]

Breakthroughs in NLP, from traditional classifiers to transformer-based models and LLMs, actively reshape social science practices by enabling scalable coding, new modes of inference, and innovative forms of human–machine collaboration, being helpful for LLM social simulations.

\end{tcolorbox}

\section{Challenges and Future Directions}
\label{sec:challenges}
 
Below, we summarize three main challenges unresolved in open-ended social simulation.

\paragraph{Data for open-ended simulation.}  Progress is constrained by the lack of datasets designed specifically for open-ended social simulation.  Most existing survey corpora contain only occasional free-text fields, leaving few matched human baselines for validating model output (\citealp[e.g., World Value Survey][]{Haerpfer2022}).  Purpose-built resources are needed in which rich demographic and contextual metadata are paired with naturally occurring open-text responses across languages and cultures.  Such data would allow researchers to compare the distribution and style of synthetic opinions with real human variation, a prerequisite for assessing whether open-ended LLM populations faithfully mirror social diversity \cite{Bender2021}.

\paragraph{Measuring alignment in infinite responses.} Open-ended social simulation makes traditional NLP metrics such as ROUGE \cite{lin-2004-rouge} or BertScore \cite{Zhang2020BERTScore} inadequate \cite{liu-etal-2023-g}, since success can no longer be defined by overlap with reference texts but by authenticity through coherence, diversity, and alignment with human behavior \cite{park2024generative}. Evaluation must therefore move from correctness to generative realism, supported by multi-dimensional frameworks. Core components include internal coherence with persona traits, preservation of meaning and style, and the measurement of generative diversity through metrics like the Sui Generis score \cite{xu2025}, which captures uniqueness beyond repetitive or stereotypical outputs. It must further include statistical measures of matching the target human distribution (\citealp[e.g.,][]{huang-etal-2024-calibrating}), not just similarity of an individual response. Ultimately, validity also requires external correspondence with real-world data and human judgments \cite{elangovan-etal-2024-considers}. This triad of coherence, diversity, and empirical realism forms a structured foundation for evaluating open-ended simulations.

\paragraph{Ethical and societal risks.} Open-ended LLM simulations raise profound ethical and epistemic concerns that extend beyond technical challenges. 
Open-ended generation can introduce distinct vulnerabilities compared with closed formats. As models are free to elaborate narratives arguments, they can produce plausible but unverifiable claims that later appear as emergent “public opinion”.
Such unconstrained outputs make it harder to audit provenance, enabling subtle bias amplification or the spread of fabricated evidence within simulated debates. 
Beyond accuracy and fairness, simulations create unprecedented risks of manipulation: coordinated populations of artificial agents could be deployed to manufacture the illusion of consensus, shaping public opinion or undermining democratic processes \cite{goldstein2023generativelanguagemodelsautomated}. Finally, relying on real-world data for persona construction introduces privacy, consent, and accountability dilemmas, with blurred lines of responsibility for harmful outputs and unresolved legal issues over data and content ownership \cite{JMLR:v24:23-0569}.

\paragraph{Future directions.} Addressing these challenges calls for several key directions: 
(i) Develop comprehensive benchmarks that pair human and synthetic open-ended responses.
(ii) Move beyond surface-level similarity metrics toward multidimensional evaluation frameworks that 
capture coherence, diversity, empirical realism and contextual appropriateness of real human responses. 
(iii) Design bias-mitigation strategies that operate during generation, not only in post-hoc detection.
(iv) Establish transparent, auditable, and participatory governance structures that guide data collection, model deployment, and evaluation practices.

\section{Conclusion}

In this position paper, we stated our position on advocating the use of open-ended questions and responses in LLM-based social simulations.  
Building on survey research traditions, we showed why and how methods for eliciting, coding, and analyzing free-text responses can guide both model design and evaluation.  
Open-ended simulation enables richer, more diverse synthetic populations and strengthens the bridge between NLP and social science.  
Future work should turn these links into practical frameworks that combine rigorous social science methodology with advances in LLMs.

\section{Limitations}

While our goal is to spark dialogue on the role of open-ended generation in LLM-based social simulation, several constraints of this position paper should be acknowledged.

\paragraph{Conceptual focus.}
Our contribution is primarily conceptual: we outline opportunities and challenges but offer no large-scale experiments or new benchmarks. The arguments rest on a synthesis of recent literature and the authors’ professional experience, serving as a call for empirical studies that can validate our claims.

\paragraph{Non-exhaustive literature review.}
The review of related work is necessarily non-exhaustive, as we are a position paper, the main focus of our paper is to advocate for the awareness of this research field with examples from representative recent papers. We focus on representative papers at the intersection of NLP and survey methodology \cite{pmlr-v235-eckman24a,sen2025connecting}, extending analyses such as \citet{anthis2025position,li2025positionsimulatingsocietyrequires} and referring readers to broader surveys like \citet{ma2024potential}, \citet{sen-etal-2025-missing} and \citet{karamolegkou2025nlpsocialgoodsurvey}. Our intent is to illustrate possibilities rather than to catalog the entire field.

\paragraph{Domain boundaries.}
We confine our discussion to the domain of LLM-based social simulation; open-ended methods for other NLP tasks or broader computational social science settings remain outside our scope. This deliberate focus keeps the argument anchored on how open-ended generation can uniquely enrich social simulation, which is the core position of our paper.

\section{Ethical Considerations}

Beyond the ethical risks regarding open-ended generation in social simulations summarized in §\ref{sec:challenges}, we highlight several broader ethical and epistemic concerns relevant to LLM simulation research, along with our own statement of the use of AI assistance.

\paragraph{Content bias of LLM responses.} Recent studies of LLM-based simulations find systematic biases in model outputs, including political left-leaning tendencies and culturally WEIRD (Western, Educated, Industrialized, Rich, Democratic) framings (\citealp[e.g.,][]{argyle-etal-2023,santurkar2023whose,cao-etal-2023-assessing,durmus2024towards,ma-etal-2025-algorithmic,vonderheyde2024vox}). These representation biases risk distorting social simulations by reproducing stereotypes rather than representing diverse populations. Importantly, most of these findings stem from closed-ended setups; whether open-ended responses amplify, mitigate, or simply reshape these biases remains an open question. A recent study also shows LLM generated survey responses are more positive than human responses \cite{Zhang2025Generative}, further questioning the validity of the LLM simulations. Future work must therefore investigate how to systematically detect and reduce representational disortions in LLM simulations, for example through bias-sensitive evaluation protocols or comparative grounding in human data.

\paragraph{Human-like validity bias.} Human survey responses face well-known validity challenges such as satisficing, order effects, and fraudulent answering \cite{10.1093/poq/nfp031,BLESS2010319,Hamby2016,10.1093/oep/gpw022,gesis-ssoar-97862}. With the rise of automated fraud in online surveys, distinguishing genuine responses from fabricated ones has become increasingly challenging \cite{10.3389/frma.2024.1432774}. In simulated settings, LLMs may produce fluent but strategically biased or shallowly reasoned outputs that mimic variability without genuine experience.
Developing validity checks, analogous to quality control in survey methodology, will be essential to prevent mistaking coherence for authenticity in simulated data.

\paragraph{Epistemic risks of ``silicon samples''.}
A final concern are the broader epistemic risks of substituting simulated agents for human participants. As noted by \citet{cummins2025threatanalyticflexibilityusing}, the danger lies in building an emerging literature that reflects methodological artifacts more than substantive social phenomena. Open-ended simulations, while rich in variability, are especially prone to analytic flexibility: coding choices, prompt designs, or sampling decisions can dramatically shape findings. More specifically, whether LLMs fit within qualitative ways of ``knowing'' as humans, remains unclear~\cite{Kapania2025}. Without transparent methodological standards, the field risks generating unreplicable insights that obscure rather than illuminate human social behavior. To avoid ``silicon-only'' social science, future work must couple LLM simulations with rigorous validation 
and clear reporting of analytic decisions.

\paragraph{Use of AI Assistance.} The authors acknowledge the use of ChatGPT (GPT-5) exclusively to paraphrase and refine the text in the final manuscript.

\section*{Acknowledgments}
We thank the members of the SODA Lab and MaiNLP Lab at LMU Munich, as well as the members of the Chair of Data Science for the Social and Economic Sciences at the University of Mannheim, for their constructive feedback. 
This work originated from discussions during BM and YC’s visit to the University of Copenhagen, kindly hosted by DH and supported by travel grants from the Danish Data Science Academy (DDSA). 
BP is supported by ERC Consolidator Grant DIALECT (101043235). 

\bibliography{custom}

\appendix
\label{sec:appendix}

\section{Detailed Use-Cases on Open-Ends in Social Simulations}
\label{sec:datailed use-cases}

In this section, we showcase a few detailed applications as use-cases for applying the open-ends in LLM social simulations from opinion polling and civic processes to focus groups, debates, behavioral experiments, and then to improving the models themselves. These applications provide low-cost, scalable, and ethically safe testbeds for studying human-like responses and collective dynamics, opening new avenues for both exploratory and policy-relevant research.

\textbf{Populations and Opinion Simulation.} LLMs can be deployed as synthetic respondents to survey or poll questions, generating answers as if they were drawn from diverse human populations \cite{cao-etal-2025-specializing, liu2025beyond, gudino2024large}. 
Over the past years, a lot of works have been focusing on using LLMs to simulate and reflect public opinions, such as simulating the US subpopulations (\citealp[e.g.,][]{santurkar2023whose}), German subpopulations (\citealp[e.g.,][]{vonderheyde2024vox}) (see a comprehensive survey in \citealp[][]{ma2024potential}). 
By conditioning on demographic, ideological, or cultural backstories, models can approximate a “silicon sample” that mirrors subgroup-specific opinion distributions. For example, an LLM might be asked to respond as a working-class urban youth or a rural conservative senior, yielding distributions of attitudes across these free-form segments \cite{argyle-etal-2023}. 
Applying open-ends opens possibilities for close, cost-effective, rapid opinion polling and exploratory pilot studies, enabling researchers and journalists to track emerging trends at the group level.  
Here, free-form answers in public opinions capture not only preferences but the rationale and nuance behind them, which are the elements that closed polls usually miss.

\textbf{Democratic Proxy Citizens.} Another emerging application positions LLMs as proxy citizens: digital stand-ins trained or prompted to represent individual or group preferences in civic processes. By conditioning on survey data, demographic profiles, or prior expressed views, these LLMs can generate plausible responses to unasked questions, forecast voting behavior on novel proposals, or fill participation gaps in deliberative settings. 
For example, \citet{karanjai2025synthesizing} propose a role-creation framework that combines demographics and personality traits to synthesize open-ended public opinion, improving alignment with real survey responses. 
This approach aims to provide policymakers with a more dynamic and granular proxy for public preference, augmenting traditional feedback mechanisms without the need for constant, exhaustive surveys. The open-ended format is crucial, as it allows these proxies to articulate evolving concerns or nuanced justifications that fixed-choice methods cannot anticipate, offering a richer understanding of citizen preferences.

\textbf{Qualitative Interviews.} 
LLMs can be deployed as synthetic participants or interviewers in qualitative research, such as in-depth interviews and focus group simulations \cite{mellor2023interview,Joongi2024,wuttke-etal-2025-ai,kim-etal-2025-llm-interviewer}. 
By conditioning on specific social roles or demographic profiles, 
researchers can elicit free-form conversational narratives. 
In multi-agent setups, these role-conditioned LLMs can interact, producing simulated dialogues that reveal discursive patterns and emergent group themes. 
For instance, \citet{slumbers2025using} demonstrate how LLMs can emulate participants in the ``Port of Mars'' collective risk social dilemma game, exhibiting human-like cooperation, communication, and leadership in open-ended exchanges. This offers a scalable, low-cost tool for piloting interview guides and exploratory thematic analysis, enabling researchers to anticipate dynamics before committing to resource-intensive fieldwork. 
Though still with limitations, as shown in a recent study \cite{Kapania2025}, open-ended prompting serves as a framework to systematically explore and strengthen these abilities.

\textbf{Deliberation and Debate Simulations.} 
Open-ended prompting lets LLMs generate arguments, counterarguments, and rhetorical moves that cannot be pre-listed, creating richer deliberative spaces than closed-choice setups. Though with limitations that they cannot fully capture human nuance in debating \cite{zhang2024llmsbeathumansdebating,Flamino2025}, LLMs are a promising tool for modeling and exploring complex communicative interactions, especially in multi-agent scenarios \cite{estornell2024multillm,liang-etal-2024-encouraging,Ashkinaze2025}.  
By staging multi-agent deliberations around contested policies, scholars can analyze argumentation dynamics, persuasion strategies, and framing effects without convening live participants. For instance, \citet{taubenfeld-etal-2024-systematic} show that while LLM agents can sustain partisan debates on salient political issues. 
This provides a controlled, low-cost environment to explore argumentative repertoires and potential pathways to consensus, offering researchers a tool to anticipate dynamics before engaging real-world participants.

\textbf{Multi-Domain Experiments and Behavior Trials.} 
LLMs can serve as synthetic participants in behavioral experiments where the open-ended format is crucial. 
LLMs can also serve as synthetic participants in controlled behavioral experiments where the open-ended format is crucial.
Rather than merely making a pre-defined choice, models can provide the reasoning and strategies behind their actions in free-form text. This provides a rich layer of qualitative data, allowing researchers to analyze not just what decision was made, but the simulated cognitive process that led to it. 
This method is being applied across diverse fields. In  psychology, for instance, LLMs generate nuanced judgments that can align with human participants \cite{Cui2025,feuerriegel2025nlp}. In economics, they can act as interactive agents in complex market games, with their open-ended negotiations revealing emergent strategies and behavioral biases \cite{brand2023using, horton2023large}. 
In this way, by analyzing these detailed, unscripted responses, LLM-based trials offer an open-ended space for testing theories, exploring counterfactuals, and probing the boundaries of human-like behavior before moving into real-world studies.

\textbf{Social Media Use-cases.} 
Besides augmenting surveys and behavioral experiments, another promising application of LLM simulations are large-scale simulations of human behavior on digital platforms, e.g., social media websites. Initial studies suggest that these simulations can offer a powerful tool for designing prototypes of new platforms~\cite{park_Social_Simulacra}, e.g., designing the rules for a new subreddit, or exploratory studies of the impact of potential interventions of exiting platforms, e.g., the impact of introducing a new rule on subreddit toxicity levels. While these social media or platform simulations are not free from challenges due to the limitations of current LLM technology~\cite{anthis2025position,nudo2025generative}, they offer many potential benefits to traditional computational simulation techniques like agent-based models. 

Some example simulations include \textit{SimReddit}, a system for developing Reddit-like social media prototypes~\cite{park_Social_Simulacra} and the comparison of different recommendation algorithms for reducing polarization~\cite{tornberg2023simulating}. We note that a majority of these simulations involve content-based social media platforms; while LLM-powered agents are also involved in engagement-based behaviors, e.g., liking and reporting, one of the main utilities of using LLMs instead of rule-based agents is their ability to generate `human-like' text (ad potentially images and videos). Therefore, for this use case, it is imperative to study the open-ended simulation abilities of LLMs. Indeed there are still many open questions about how to best validate LLM-based social media simulations, given the complex, multifaceted behavior of these agents and resultant macro-outcomes of a simulation. To validate whether LLMs are good proxies of humans in social media simulations, a researcher would compare social media posts generated by LLM agents to those authored by humans. However, there are no clear evaluation standard for this comparison. One could use computational techniques (e.g., embedding-based similarity measures), qualitative techniques (human annotators being asked to differentiate generated and real posts, as in \citealp[][]{park_Social_Simulacra}), or a combination.

In summary, social media or digital simulations are a promising use case for LLMs; but it is also a use case that primarily depends on faithful and `human-like' open-text generations from them, making it all the more essential for NLP researchers to develop standards for evaluating open-text generations of LLMs in social simulations.

\textbf{Model Training and Evaluation.}  
The benefits of open-ended responses extend beyond social science applications and have direct implications for NLP modeling. The relationship between NLP and social science is bidirectional: while NLP offers powerful tools for advancing social good \cite{karamolegkou2025nlpsocialgoodsurvey}, insights from social science can in turn strengthen NLP modeling and evaluation practices \cite{pmlr-v235-eckman24a,sen2025connecting}. Surveys, a cornerstone of social science, are increasingly employed in NLP research to elicit both qualitative and quantitative human feedback (\citealp[e.g.,][]{argyle-etal-2023,santurkar2023whose,cao-etal-2023-assessing}). Importantly, recent advances such as reinforcement learning from human feedback (RLHF), a central method for fine-tuning large language models, rely heavily on human preference data \cite{Christiano2017,leike2018scalableagentalignmentreward,ouyang2022training,kaufmann2024survey}. In this context, open-ended responses offer a largely untapped resource. Their richness and variation provide nuanced human judgments that go beyond binary or scalar preference signals. For model training, they can expose subtle distinctions in reasoning and articulation that help align LLMs with human-like perspectives. For evaluation, they enable more human-centric assessments, offering benchmarks that capture complexity and diversity rather than reducing performance to rigid accuracy scores. In short, integrating open-ended responses into NLP pipelines can improve both the fidelity and the validity of model alignment.

\section{Social Science Practice: Coding Scheme}
\label{sec:coding-scheme}
\begin{figure}[htbp]
    \centering
    \includegraphics[width=1\linewidth]{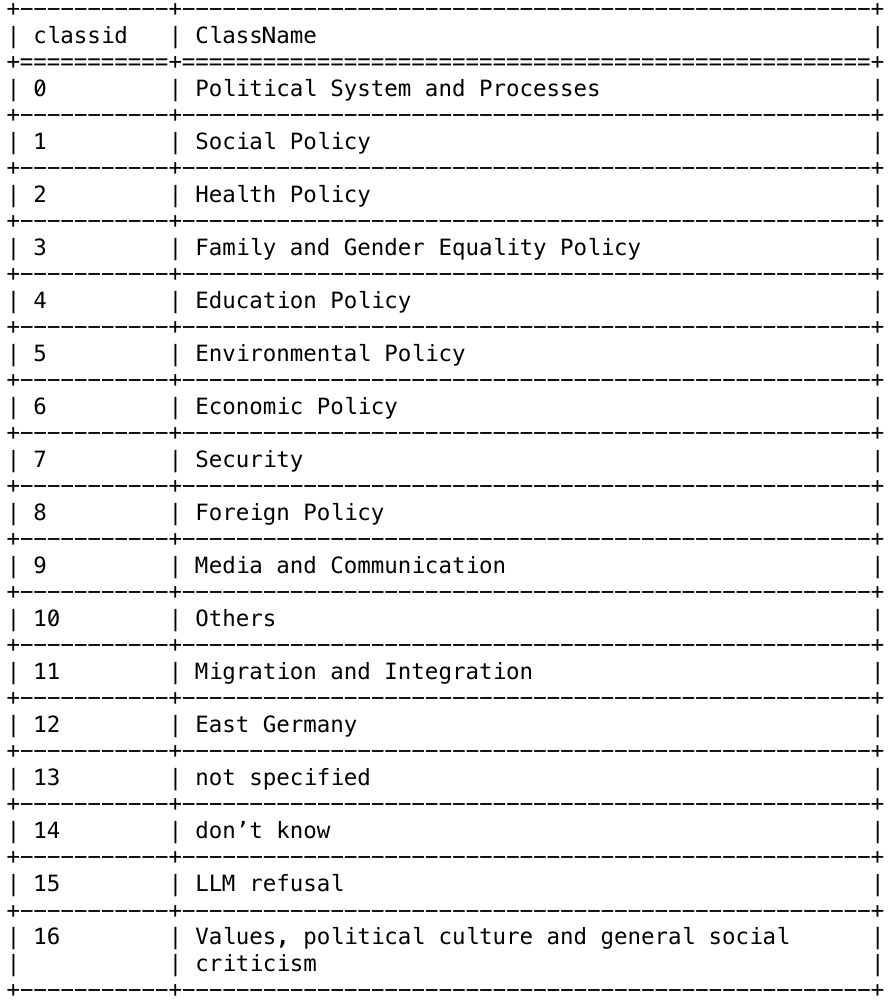}
    \caption{An example coding scheme modified for coding open-ended LLM responses to a question ``In your opinion, what is the most important question facing Germany today?''. 
    The human-coders (or LLM-Coders) are instructed to categorize the open-ended responses into the listed classes.}
    \label{fig:coding-scheme}
\end{figure}

Figure \ref{fig:coding-scheme} shows an example coding scheme of 16 categories used for coding the LLM responses in a Germany-based recent simulation study \cite{ma-etal-2025-algorithmic}. This is a common practice in the social science to code the open-ended survey responses into defined categories for better measurability.

\end{document}